%% file: Article.tex
\documentclass[11pt,a4paper]{article}
\usepackage{times,latexsym}
\usepackage{amsmath}
\usepackage{amssymb} 
\usepackage{mathtools} 
\usepackage{svg} 
\usepackage[font=small]{caption} 
\usepackage{multirow} 
\usepackage{url}
\usepackage[T1]{fontenc}

\usepackage[acceptedWithA]{tacl2021v1}

\usepackage{xspace,mfirstuc,tabulary}

\newif\iftaclinstructions
\taclinstructionsfalse 
\iftaclinstructions
\renewcommand{\confidential}{}
\renewcommand{\anonsubtext}{(No author info supplied here, for consistency with
TACL-submission anonymization requirements)}
\newcommand{\instr}
\fi

\iftaclpubformat 

\else

\fi

\newcommand{\R}{\mathbb{R}}
\newcommand{\M}[1]{\mathbf{#1}}
\newcommand{\T}[1]{\mathbf{#1}^\top}
\newcommand{\MT}[1]{\T{\M{#1}}}

\title{Query-Focused Extractive Summarization for Sentiment Explanation
 \thanks{This work was supported by Mitacs through the Mitacs Accelerate program.}
}

\author{
  Ahmed Moubtahij$^{1,2}$ 
  Sylvie Ratté$^2$ 
  Yazid Attabi$^1$
  \and
  Maxime Dumas$^1$
  \\
  \ \\
  $^1$Croesus Lab
  \\
  Laval, Québec, Canada
  \ \\
  \\
    $^2$Software Engineering and IT Dept.
  \\
   École de Technologie Supérieure
  \\
   Montreal, Canada
}

\date{}

\begin{document}
\maketitle

\begin{abstract}\footnote{This study originates from the master's thesis of the first author (2022-2023), which was completed before the advent of large language models (LLMs) like ChatGPT.}
  \input{Sections/Abstract}
\end{abstract}

\section{Introduction} 
\input{Sections/Introduction}

\section{Related Work}
\input{Sections/RelatedWork}

\section{Methodology}
\input{Sections/ProposedApproaches}

\section{Experiments}\label{EXP}
\input{Sections/Experiments}

\section{Results and discussion}\label{ResAndDisc}
\input{Sections/ResultsAndDiscussion}

\section{Conclusion}
\input{Sections/Conclusion}




\bibliography{tacl2021}
\bibliographystyle{acl_natbib}

\end{document}

%% file: Sections/Abstract.tex
Constructive analysis of feedback from clients often requires determining the cause of their sentiment from a substantial amount of text documents. To assist and improve the productivity of such endeavors, we leverage the task of Query-Focused Summarization (QFS). Models of this task are often impeded by the linguistic dissonance between the query and the source documents. We propose and substantiate a multi-bias framework to help bridge this gap at a domain-agnostic, generic level; we then formulate specialized approaches for the problem of sentiment explanation through sentiment-based biases and query expansion. We achieve experimental results outperforming baseline models on a real-world proprietary sentiment-aware QFS dataset.

%% file: Sections/Introduction.tex
Sentiment analysis is the Natural Language Processing (NLP) task of predicting the affective state of a text passage. It is generally useful for applications concerned with feedback analysis of experiences (e.g., products, events, or services). However, simply being aware of the sentiment does not enable improvement of the experience; this purpose requires knowledge of the specific causes and features related to the sentiment.

Given a multitude of documents, a sentiment of interest (e.g., negative or positive), and a query regarding the targeted entities (e.g., a specific product, date, or location), our main objective is to provide an informative summary of the input documents that explains the cause(s) of the queried sentiment. This goal falls under a constrained QFS task, which we term Explicative Sentiment Summarization (ESS). See Figure \ref{fig:SAMBTR_fig} for a depiction of this process.

Compared to the Question Answering task's factoid outputs, the QFS task is motivated by more complex and contextually rich responses. It is thus a more appropriate parent task for ESS, which consists of elaborating on the cause(s) of the queried sentiment. The problem space of ESS is marginally akin to that of the Aspect-Based Sentiment Analysis (ABSA) task. ABSA associates sentiments with specific aspects (categories, features, or topics). Such aspects are predefined or extracted by a pipeline component, and the sentiment of each is a prediction objective. ESS concerns use cases where the target sentiment is prior knowledge and is thus an input item. Leveraging the latter allows simplifications such as computing the strength of the targeted sentiment for each text passage, thus inherently circumventing aspect identification. Additionally, ABSA produces sentiment associations for each aspect, whereas ESS outputs a natural language summary explaining the cause of the queried sentiment.

A common shortcoming of the QFS task and its proposed models is the putative gap between the source text and the input query in terms of \textit{Language Register} (LR, formality level) and \textit{Information Content} (IC, from Shannon's Information Theory). An LR gap occurs when, for example, a colloquial query formulation addresses source text written in formal style or in domain-specific terminology. An IC gap is typically incurred by the generic semantic coverage of short queries in relation to the specific semantics in detailed source text passages.

Our following contributions first address this issue at a generic level, then at a specialized level for our purpose of sentiment explanation:

\begin{enumerate}

    \item We introduce the \textit{Compound Bias-Focused Summarization (CBFS)} (\ref{CBFS}) framework to improve the chances of aligning the user's intent with arbitrary and possibly heterogeneous language registers in source documents by supporting multiple query formulations;

    \item We concretize the CBFS framework with our \textit{Multi-Bias TextRank} (MBTR) (\ref{MBTR}) model and its \textit{Information Content Regularization} (\ref{ICR}) which guides the QFS process towards the desired level of specificity;
    
    \item We introduce the \textit{Explicative Sentiment Summarization (ESS)} task, (\ref{ESS}) which specializes the QFS task by leveraging prior knowledge in a sentiment explanation setting;
    
    \item We substantiate the ESS task with sentiment-based bias computation (\ref{SB}) and query expansion (\ref{FSQE}).
\end{enumerate}

%% file: Sections/RelatedWork.tex
The following is an overview of the literature relevant to our task and contributions, spanning works in query-focused extractive summarization and query expansion.

\subsection{Query-Focused Extractive Summarization}\label{MDQFES}

The NLP task of automatic summarization aims to compress a document or collection of documents into a salient and concise summary. \citet{jones_automatic_1998} introduces three context factors concerned with automatic summarization and its evaluation: the nature of the input text (e.g., its domain and structure); the nature of the output summary; the purpose of the summary. \citet{ter_hoeve_what_2022}, who ground their work in that of \citet{jones_automatic_1998}, advocate for the usefulness of a summary concerning the user's needs. They report that the purpose factors receive the least attention from works in automatic summarization, barring specializations which consider the audience and the situation. Among the latter is the QFS task, of which the expected output is a summary of the input document(s) that focuses on the query.

Automatic summarization can be achieved either by a semantic abstraction of the source text's salient information, or by a verbatim extraction of it.

While human-level summarization is abstractive, in practice, recent works (\citet{ladhak_tracing_2022}; \citet{ladhak-etal-2022-faithful}; \citet{balachandran_correcting_2022}; \citet{fischer_measuring_2022}) are still attempting to solve text generation errors such as factuality and hallucination. These shortcomings make Query-Focused Abstractive Summarization (QFAS) models currently unreliable in applications with tangible stakes.

Extractive summarization selects and concatenates salient text spans. This approach potentially hinders the cohesion of the summary as a whole. Indeed, text cohesion is a generally desired attribute and yet one of the most common error types in extractive summaries (\citet{kaspersson_this_2012}; \citet{smith_cohesion_2012}). However, it may be an optional attribute for critical applications prioritizing content reliability, output traceability, and fact-checking, all facilitated in Query-Focused Extractive Summarization (QFES).

Numeric representation of text is ancillary to the automatic summarization task, since it enables arithmetic transformations from the task's input space to its output space. Given the importance of pragmatics in natural language, the usefulness of such representations is greatly improved by their sensitivity to context. \textit{BERT} \cite{devlin-etal-2019-bert}, a pre-trained transformer \cite{vaswani_attention_2017} encoder-based architecture, has seen widespread use as a Pre-trained Language Model (PLM) across recent text summarization systems (\citet{liu_text_2019}; \citet{laskar_query_2020}; \citet{kazemi_biased_2020}; \citet{laskar-etal-2020-wsl}; \citet{xu-lapata-2020-coarse}; \citet{xu-lapata-2021-generating}; \citet{xu-lapata-2022-document}; \citet{laskar_domain_2022}). These models' State-Of-The-Art (SOTA) performance motivated us to adopt BERT-based models for text representation in automatic summarization.


Maximal Marginal Relevance (MMR) \cite{carbonell_use_1998} is a Multi-Document Query-Focused Extractive Summarization (MDQFES) algorithm that conjointly considers diversity and query relevance when retrieving salient passages from a collection of documents.

\citet{liu_text_2019} propose \textit{BertSum}, a BERT-based model fine-tuned for both abstractive and extractive summarization, respectively, on the \textit{XSUM} \cite{narayan_dont_2018} dataset as \textit{BertSumAbs}, and on the \textit{CNN/DailyMail} \cite{hermann_teaching_2015} dataset as \textit{BertSumExt}. \citet{laskar_query_2020} pre-train BertSum similarly to BertSumAbs, then fine-tune it on the \textit{DebatePedia} dataset \cite{nema_diversity_2017} for QFAS.

Motivated by the success of BERT contextual embeddings, \citet{kazemi_biased_2020}'s unsupervised \textit{Biased TextRank} (BTR) model represents nodes from \textit{TextRank} \cite{mihalcea_textrank_2004}, a complete graph, as \textit{SBERT} \cite{reimers_sentence-bert_2019} sentence encodings. BTR then subjects the underlying \textit{PageRank} \cite{page_pagerank_1999} centrality computation to a lower bound similarity, and to a query-bias for every sentence-node. Thereby ranking the input sentences by a conjunction of their centrality and query-bias.

\citet{xu-lapata-2020-coarse} argue that disjoining intra-document salience and query-relevance allows for separate modeling of the query and for summaries to address specific questions; this motivates their coarse-to-fine model, \textit{QuerySum}, where text passages from the input documents are sequentially processed through query-relevant retrieval, followed by evidence estimation based on the Question-Answering (QA) task, and then by centrality-based re-ranking, i.e., salience for the surrounding text passages.

\citet{laskar-etal-2020-wsl}, \citet{xu-lapata-2021-generating}, \citet{xu-lapata-2022-document} and \citet{laskar_domain_2022} propose different approaches to address the prominent issue of lack of labeled QFS datasets. 

\citet{laskar-etal-2020-wsl} opt for a distant and weakly supervised approach for generating weak (artificially generated) reference summaries from gold reference summaries through a pre-trained, RoBerta-based \cite{liu_roberta_2019} sentence-similarity model.

Assuming that generic (non-query-focused) summaries contain information on latent queries, the \textit{MARGE} model \citep{xu-lapata-2021-generating} uses selective masking to reverse-engineer proxy queries, then pairs them with input sentences scoring high on \textit{ROUGE} \cite{lin-2004-rouge} (see \ref{ROUGE}). This pairing enables weak supervision for ranking query-relevant sentences that are subsequently fed to a length-controllable QFAS model with optional user-query.

The \textit{LQSum} model \citep{xu-lapata-2022-document}, unlike MARGE, does not assume the target queries' length and content, nor does it require a development set. It achieves this by discarding the sequential query modeling approach, and replacing it with a zero-shot-capable alignment between the source tokens and discrete latent variables. The latter are expressed by a binomial distribution indicating the query relevance belief of a source token.


\subsection{Query Expansion}\label{QE}

The dissonance between query and object signals motivates the NLP task of Query Expansion (QE), which is ancillary to downstream tasks such as QA, Information Retrieval (IR), or QFS. QE generally employs techniques such as re-weighting query terms and/or augmenting them with semantically related terms (\citet{riezler_statistical_2007}; \citet{ganu_fast_2018}; \citet{zheng_bert-qe_2020}
).

\citet{riezler_statistical_2007}'s query expansion methods leverage Statistical Machine Translation (SMT) for paraphrasing and mapping to answer terms. While such back translation methods might somewhat preserve semantics, they are liable to lose the domain property of language, which disqualifies it from our need to bridge the language register gap between user-query and domain-specific documents. This particular discrepancy is observed by \citet{ganu_fast_2018} in the search feature of their accounting software, in which users employ colloquial language to query the formal and financial text in their knowledge base. They address this problem with strategies for synonym substitution and expansion to nearest neighbor-embeddings, based on vocabulary from their hand-curated proprietary dataset. Albeit a valid approach for aligning the domain of query language, crafting a problem-specific lexicon requires seldom available human resources and expertise.

\citet{zheng_bert-qe_2020} further the motivation of QE with the issue of noisy query expansion, for which they propose \textit{BERT-QE}, a three-step QE model in which initially ranked documents are: 1) re-ranked on query-relevance with a BERT model pre-trained on the \textit{MS MARCO} \cite{bajaj_ms_2018} QA dataset; 2) chunked into passages for relevance scoring with the model fine-tuned on a target dataset; 3) re-ranked based on passage document-relevance and query-relevance. \citet{zheng_bert-qe_2020}'s QE approach is restricted to IR as a downstream task by considering retrieval objects as entire documents, which does not directly accommodate our target task of QFES since it retrieves sentences.

Akin to the QE task, the Term Set Expansion (TSE) task consists of expanding members of a semantic class from a small seed set of terms. \citet{kushilevitz_two-stage_2020} propose two TSE methods based on BERT used directly as a Masked Language Model (MLM): 1) In \textit{MPB1} (MLM-Pattern-Based), seed-terms are masked in sentences in which they occur (indicative patterns), then an MLM predicts the masks in their contexts, at which point the correctly predicted masks have their next best predictions elected for query expansion; 2) \textit{MPB2} circumvents out-of-vocabulary masked terms in indicative patterns by querying similar patterns for single- and multi-token terms.

\citet{kushilevitz_two-stage_2020}'s methods leverage an MLM's vocabulary for expanding seed terms in the context of the input text, which does not require a handcrafted lexicon and helps align the source documents' language register with that of the expanded seed-terms. In our work, we need only consider seed terms as query terms to utilize these TSE methods for query expansion.

%% file: Sections/ProposedApproaches.tex
We establish a framework for combining multiple queries, concretize it with our MBTR model, then subject the latter to information content regularization. Then, we introduce the ESS task for sentiment explanation and employ corresponding techniques with reference-based query formulation, sentiment bias, and query expansion.

\subsection{Compound Bias-Focused Summarization}\label{CBFS}

To the best of our knowledge, all current QFS models consider a single input query. This design burdens the query's formulation by targeting all information of interest at various scopes of variance and depth. Presented with such a challenge, all query formats \cite{xu-lapata-2021-generating} face the following difficulties: natural language articulation must encompass the full intent; keywords circumvent the syntactic constraints of natural language at the cost of its expressive flexibility (e.g., contextual disambiguation); albeit concise, the typical brevity of a title might limit the specificity of attainable information; a composite of the latter formats allows for trade-off balancing but incurs a non-trivial choice of representation to accommodate its syntactic heterogeneity effectively.

\begin{figure*}
    \centering
    \includegraphics[keepaspectratio, width=0.75\textwidth]{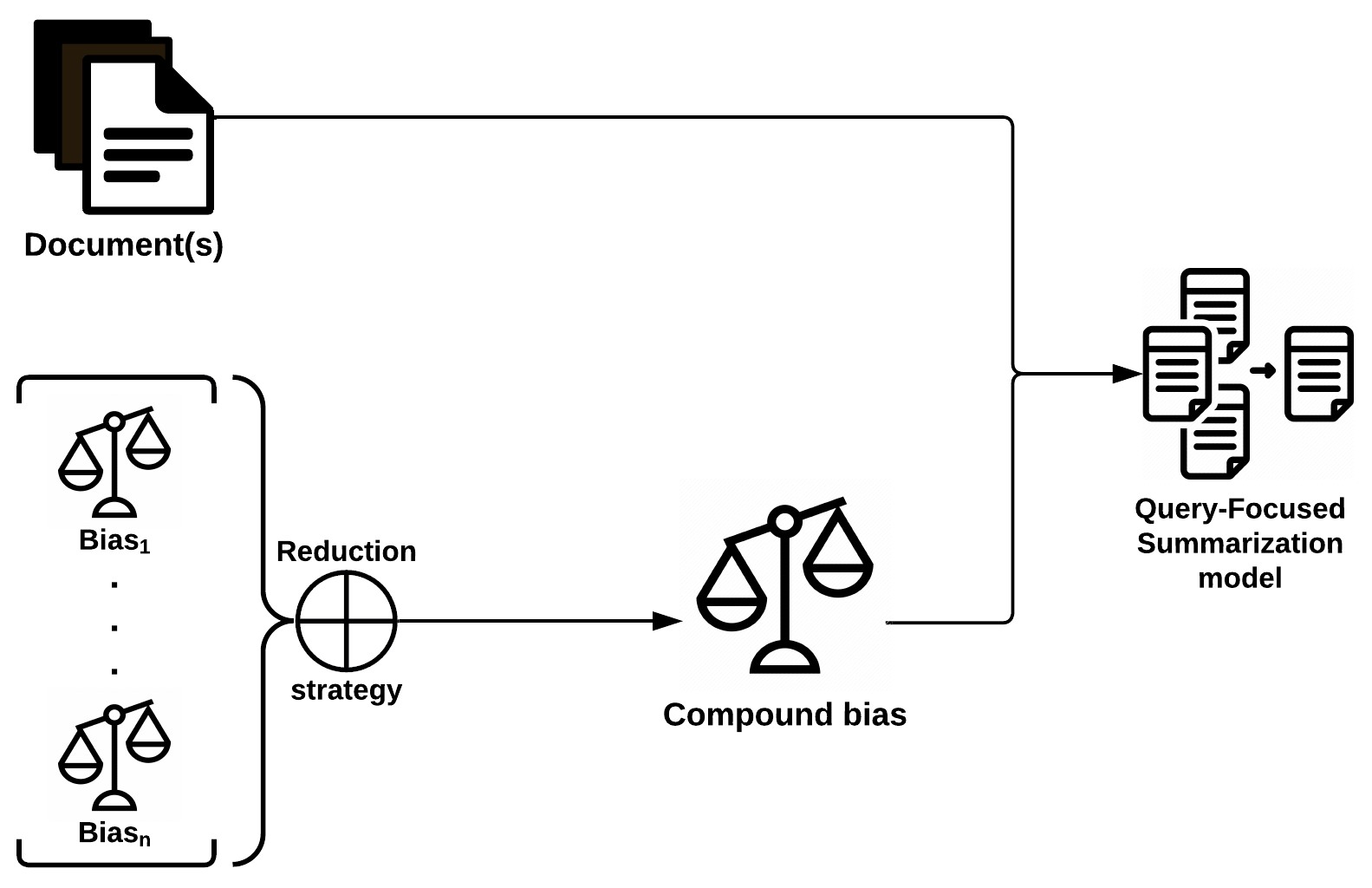}
    \captionsetup{justification=centering}
    \caption{Compound Bias-Focused Summarization framework. The contributions of multiple biases are folded into a compound bias, which is then integrated into a Query-Focused Summarization model.}
    \label{fig:CBFS_fig}
\end{figure*}

To tackle the aforementioned challenge, we propose the \textit{Compound Bias-Focused Summarization (CBFS)} framework (Figure \ref{fig:CBFS_fig}). In CBFS, the effects of multiple biases are combined through a reduction strategy\footnote{(weighted) summation, max, conjoint probabilities, median, inverse variance, etc.} and input to a QFS model. We use the term "bias" as a generalization over skewings of both query and non-query (e.g., \ref{SB}) natures. Providing multiple bias channels alleviates the burden in query formulation by partitioning the compromises mentioned above, instead of imposing them on a single query. Intuitively, this is analogous to humans reformulating questions from multiple perspectives or through various language registers for a wider coverage of their audience. Audience consideration is a heading of the advocated summarization purpose factor (\citet{jones_automatic_1998}; \citet{ter_hoeve_what_2022}).

\subsection{Multi-Bias TextRank}\label{MBTR}

Given its simplicity and flexibility, we extend \citet{kazemi_biased_2020}'s BTR model to \textit{Multi-Bias TextRank} to demonstrate the proposed CBFS framework.

Let \(n\) sentence encodings, \(d\) the embedding dimension, \(\M{b} \in \R^{d}\), \(\M{S} \in \R^{n \times d}\), \(\alpha\) a control parameter, \(\theta\) the similarity threshold and \( \widetilde{\M{A}} \in \R^{n \times n} \) a lower-bounded normalization of the weighted adjacency matrix \( \M{A} = sim(\M{S}, \M{S}) \) such that:

\begin{equation}\label{norm_adj_eq}
\widetilde{\M{A}}_{ij} =
\begin{cases}
\frac{\M{A}_{ij}}{\sum\limits_{j=1}^n \M{A}_{ij}},& \text{if} \sum\limits_{j=1}^n \M{A}_{ij} \neq 0 \text{ and } \frac{\M{A}_{ij}}{\sum\limits_{j=1}^n \M{A}_{ij}} \geq \theta\\
0,                                                & \text{otherwise}
\end{cases}
\end{equation}

Then the PageRank vector in the Biased TextRank model can be recursively computed as follows:

\begin{equation}\label{btr_eq}
\begin{split}
R = \alpha \widetilde{\M{A}} R \,+\, &(1 - \alpha) sim(\mathbf{b}, \M{S})
\end{split}
\end{equation}

Let \(q\) the number of query encodings, \(\M{B} \in \R^{q \times d}\) and \( \mu \) a normalization function such as
\(
\mu :\R^n \setminus \{\mathbf{u}: \T{1} \mathbf{u} = 0\} \to \R^n : \mathbf{u} \mapsto \mathbf{u} / (\T{1} \mathbf{u})
\). Then the PageRank vector in our Multi-Bias TextRank model is expressed as follows:

\begin{equation}\label{mbtr_eq}
\begin{split}
R = \alpha \widetilde{\M{A}} R \,+\, &(1 - \alpha) \mu\left( \bigoplus\limits_{i=1}^q sim(\mathbf{B}, \M{S})_{i^*} \right)
\end{split}
\end{equation}

We implement the \( \oplus \) reduction operator as a summation, and the similarity function \(sim: \R^{m \times k} \times \R^{n \times k} \to \R^{m \times n}\ \) as matrical cosine similarity:

\begin{equation}
sim(\M{U}, \M{V})_{ij} \coloneqq \frac{(\M{U}\MT{V})_{ij}}{||U_{i*}|| \cdot ||V_{j*}||}    
\end{equation}

where \(\coloneqq\) denotes "defined as", and the \( i* \) and \( j*\) subscripts denote a row-vector of a matrix. The \( \mu \) normalization of the cumulative bias vector scales it comparably to the centrality vector \( \widetilde{\M{A}} R \).

While a single query formulation might not effectively address a desired sentence, folding the bias vectors of multiple queries increases its relevance score. Conjointly, a low score denotes more confidence in rejecting a sentence, given the implication that none of the query formulations neighbor it in semantic space.

The PageRank recursive term, \(R\), in equations \ref{btr_eq} and \ref{mbtr_eq}, computes centrality through the repeated transformation of itself by the weighted adjacency matrix \( \widetilde{\M{A}} \). Thus, \(R\) is essentially converging towards the eigenvector of \( \widetilde{\M{A}} \) with an eigenvalue of 1, i.e., the stationary probability distribution of the salience likelihoods of each sentence. Intuitively, this process simulates the broadcasting of sentence salience throughout the TextRank graph. In other words, it iteratively amplifies the scores of sentences similar to important sentences until convergence\footnote{A set number of iterations and/or an \(\epsilon\) error tolerance.}. Once \( \widetilde{\M{A}} \)'s equilibrium distribution is sufficiently stable, the sentences associated with the top probabilities are selected as the output summary. 

\subsection{Information Content Regularization}\label{ICR}

\citet{amigo_information_2022} call attention to the formal properties of text embeddings, based on the notion of Information Content (IC) from Shannon's Information Theory. One such property is the correspondence of IC with the vector norm of a text unit's embedding. We leverage this feature to disfavor candidate sentences by their distance from the targeted level of specificity.

Let \( \M{G} \in \R^{m \times d} \) a matrix of \(m\) sentence-encodings from a guiding example summary, and \( \mathbf{\Delta_{\text{IC}}} \in \R^n \) the observed-to-target IC distances:

\begin{equation}\label{delta_ic_eq}
\mathbf{\Delta_{\text{IC}}}_i \coloneqq |\, ||\M{S}_{i*}|| - avg(\, (||G_{j*}||)_j\, )  \,|
\end{equation}

where \( avg: \R^n \to \R \) denotes a statistical average, which we define as the arithmetic mean \( avg(\mathbf{u}) \coloneqq  \bar{\mathbf{u}} \). Then, with \( \beta \) as a control parameter\footnote{Note that \(\text{BTR} \equiv \text{MBTR}|_{q=1, \beta=0}\)}, we penalize every bias vector \(sim(\mathbf{B}, \M{S})_{i^*}\) in Equation \ref{mbtr_eq} by its distance from the target IC (Equation \ref{delta_ic_eq}):

\begin{equation}\label{icr_eq}
\begin{split}
R =\, &\alpha \widetilde{\M{A}} R \,+\\
    &(1 - \alpha) \mu\left( \bigoplus\limits_{i=1}^q (sim(\mathbf{B}, \M{S})_{i^*} - \beta \mathbf{\Delta_{\text{IC}}}) \right)
\end{split}
\end{equation}

The sentences associated with \(\mathbf{G}\) can be provided by application-specific prior knowledge (see \ref{MBTR_ICR_QE}), in which case the target IC, i.e., \( avg(\, (||G_{j*}||)_j \,) \) is embedded in the system, or by a user's example text to guide the desired level of specificity.


\subsection{Explicative Sentiment Summarization}\label{ESS}

For sentiment explanation, we can disregard open-domain queries and specialize the QFS task for biases and queries that align with this objective. Additionally, we can leverage the prior knowledge of queries in a sentiment explanation setting. In the following sections, we introduce the task of \textit{Explicative Sentiment Summarization (ESS)}.

\subsubsection{Reference-based Query Formulation}\label{RQF}

For any sentiment-aware QFS dataset, its summary references are expected to explain the queried sentiments. We leverage this expectation to dispense users of query formulation by automating it in the ESS model, thus reducing the user query's burden to merely mentioning the specific entities of interest, such as product names or dates, which can then be appended to the automated query or considered a separate query as per \ref{CBFS}.

A simple heuristic for automating query formulation in an ESS setting would be selecting the \textit{Frequent Reference-Words (FRW)} or \textit{Frequent Reference-Phrases (FRP)} from the development split of the ESS dataset. This approach has the advantage of embedding common answer signals directly into the QFS bias.

\subsubsection{Sentiment Bias}\label{SB}

Unlike the QFS task, ESS can make assumptions about the query, such as the user's prior knowledge regarding the sentiment of interest. This allows an ESS model to adapt its query-relevance computation consequently.

Sentiment classifiers are trained to predict the perceived polarity of a text passage. The use case of sentiment explanation assumes prior knowledge of the sentiment of interest; we can thus utilize the probabilistic confidence in this sentiment for every input sentence to construct a \textit{sentiment bias vector}. However, the latter is potentially insufficient for the ESS task since it does not encode information regarding the targeted entities (e.g., product name) and should thus be used in combination with complementary query-biases (\ref{CBFS}), as exemplified in Figure \ref{fig:SAMBTR_fig}.

This ESS-specific approach demonstrates a novel bias method that contrasts with the conventional query-sentence similarity computation in QFS.

\subsubsection{Sentiment-based Query Expansion}\label{FSQE}

In addition to enabling a sentiment bias vector (\ref{SB}), the prior knowledge in ESS can also be utilized for sentiment-based query expansion.

We propose using a hyperparameter pair of small sentiment phrases to select from for expansion, for example, "excellent service" and "poor experience". The suggested brevity is motivated by its correlation with low Information Content (\ref{ICR}), i.e., less specificity, which should broaden the reach for expansion in semantic space. We use phrases as text units instead of words to leverage the collocational properties of PLMs and thus enhance representation in semantic space.

Given an input sentiment, the ESS system: 1) selects the corresponding integrated sentiment phrase; 2) decomposes the input document(s) into phrases (see \ref{SAMBTR}); 3) retrieves the top \(K\) document-phrases with the most cosine-similar encodings to the sentiment phrases; encodings in this step are produced with an asymmetric semantic search encoder \footnote{\url{https://www.sbert.net/examples/applications/semantic-search/README.html}} given the brevity of the sentiment-phrase. See Figure \ref{fig:SAMBTR_fig} for a depiction of this process.

This QE method does not require an external lexicon or knowledge base and inherently circumvents the typical linguistic dissonance between the query and the source document(s).

%% file: Sections/Experiments.tex
We present the used dataset and the evaluation metric, then apply our proposed methods in two main experiments: \textit{MBTR with query expansion}, which requires a development set, and \textit{MBTR with sentiment}, which does not.

\subsection{Dataset}\label{dataset}

We use a proprietary ESS dataset of which only metadata is disclosable. This dataset spans 950 ESS units, each containing:

\begin{itemize}
    \item the name of the targeted entity
    
    \item the sentiment of interest
    
    \item 1 to 576 documents with a mean of 17 and variance of 38, with each document spanning 2 to 771 sentences with a mean of 24 and variance of 36
    
    \item a single-sentence abstractive reference summary explaining the sentiment
\end{itemize}

We conduct experiments using 75\% of examples as a development set, and 25\% as a test set. 

\subsection{Evaluation Metric for Automatic Summarization}\label{ROUGE}

We use the Recall-Oriented Understudy for Gisting Evaluation (ROUGE) \cite{lin-2004-rouge} metric as it is the de-facto standard in automatic summarization. ROUGE varies strategies to quantify the n-gram overlap of the output text with its reference(s). Our ESS dataset presents single-sentence summaries of multiple documents; \citet{lin-2004-rouge} reports the ROUGE-\{1, L, SU4, SU9\} variants as most correlating with human judgment in the \textit{problem space of short summaries}. However, \citet{owczarzak_assessment_2012} advocate for ROUGE-2-R, \citet{rankel_decade_2013} for ROUGE-\{3, 4\}, and \citet{graham_re-evaluating_2015} for ROUGE-2-P.

Given the above discordance, we heuristically elect ROUGE-SU4 by the criterion of top variance through numerous experimental runs on our dataset, hypothesizing that high variance denotes reactivity to summary quality and low variance insensitivity to it; thus, we report \textit{ROUGE-SU4}. We find that its F1 score is also reported in recent works (\citet{xu-lapata-2020-coarse}; \citet{xu-lapata-2021-generating}; \citet{xu-lapata-2022-document}; \citet{laskar_domain_2022}) in combination with the F1 scores of \textit{ROUGE-1, ROUGE-2} and \textit{ROUGE-L} (\citet{laskar_query_2020}; \citet{kazemi_biased_2020}), which we also report using the \textit{pythonrouge}\footnote{\url{https://github.com/tagucci/pythonrouge}} implementation.

\subsection{Multi-Bias TextRank with Query Expansion}\label{MBTR_ICR_QE}

We use the NLTK \cite{bird2009natural} library to decompose the input documents into sentences, and an SBERT\footnote{\href{https://huggingface.co/sentence-transformers/xlm-r-distilroberta-base-paraphrase-v1}{xlm-r-distilroberta-base-paraphrase-v1}} encoder to represent them and the following expanded queries in \(\text{MBTR}|_{\alpha=\beta=0.1}\) (Equation \ref{icr_eq}):

\begin{enumerate}
    \item \textbf{FRW-MPB2}: we construct an FRW query with the top 20 frequent non-stopwords from the development set, then expand it with MPB2 (\ref{QE}), using its authors' \cite{kushilevitz_two-stage_2020} reported hyperparameters.
    \item \textbf{FRP-MPB2}: we redefine text units in FRW-MPB2 as noun phrases, which we obtain using the spaCy \cite{ines_montani_2020_4091419} library's noun chunks feature\footnote{\url{https://spacy.io/usage/linguistic-features\#noun-chunks}}.
    \item \textbf{FRP-BTR}: we expand the FRP query using BTR \cite{kazemi_biased_2020} with phrases as text units\footnote{\url{https://github.com/DerwenAI/pytextrank}}, then re-rank its output by descending frequency in the input documents and retrieve the top 20 phrases.
\end{enumerate}

Before concatenating the individual terms (words or phrases) for each of the above query expansions, we remove duplicates, terms entirely composed of stopwords, and mentions of specific entities such as dates or organization names – using spaCy's NER\footnote{\url{https://spacy.io/usage/linguistic-features\#named-entities}} feature – to avoid spurious skewing towards a subset of the input sentences. We preserve \citet{kazemi_biased_2020}'s recommended \(\theta\!=\!0.65\) for the similarity threshold (Equation \ref{norm_adj_eq}) in all (M)BTR experiments.

The FRW-MPB2 + FRP-MPB2 + FRP-BTR query combination will hereafter be referred to as \textit{Expanded Reference-Terms (ERT)}.

In the ESS task, we prepend the targeted entity's name to each query before and after expansion. Doing so produces deliberate skewing towards entity-relevant sentences. Additionally, we construct the \(\M{G}\) encodings matrix in Equation \ref{delta_ic_eq} from reference sentences in the development set.

\subsection{Sentiment-aware Multi-Bias TextRank}\label{SAMBTR}

\begin{figure*}
    \centering
    \includegraphics[keepaspectratio, width=0.95\textwidth]{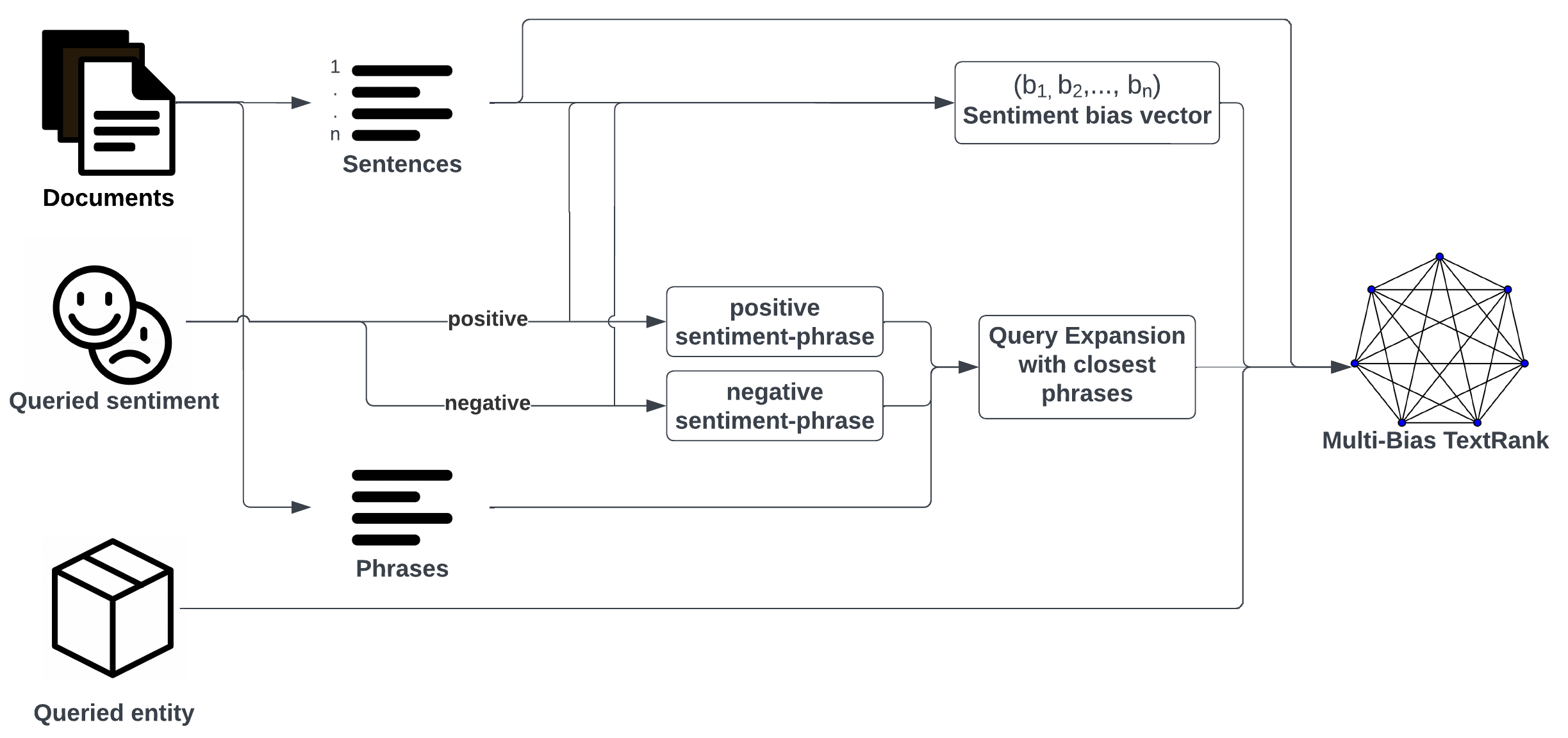}
    \captionsetup{justification=centering}
    \caption{Explicative Sentiment Summarization system: integration of sentiment-based query expansion and sentiment bias into Multi-Bias TextRank.}
    \label{fig:SAMBTR_fig}
\end{figure*}

Given input documents, a queried entity and sentiment, Figure \ref{fig:SAMBTR_fig} depicts the following process:

\begin{enumerate}
    \item We use a sentiment classifier to predict the probability of the given sentiment for every input sentence, thus producing a \textit{sentiment bias vector}.

    \item We select the sentiment-corresponding query from a hyperparameter pair of sentiment phrases, then expand it to its top \(K\) most cosine-similar document phrases\footnote{We use K=30} in the space of an asymmetric semantic search encoder\footnote{\href{https://huggingface.co/sentence-transformers/msmarco-distilbert-base-v4}{msmarco-distilbert-base-v4}}. The resulting expanded queries are prepended with the queried entity.
    
    \item We combine the sentiment bias vector with the expanded queries' bias vectors in \(\text{MBTR}|_{\alpha=0.1, \beta=0.2}\) (Equation \ref{icr_eq}).
\end{enumerate}

In the second step above, phrases are noun phrases (NP) and verb phrases (VP). NPs are extracted with spaCy's noun chunking feature, as declared in \ref{MBTR_ICR_QE}. We specialize VP patterns for the ESS task using spaCy's rule-based matching\footnote{\url{https://spacy.io/usage/rule-based-matching}} such as:

\begin{verbatim}
    0. vp_pattern = [
    1. {},
    2. {'POS': 'AUX', 'OP': '?'},
    3. {'DEP': 'neg', 'OP': '?'},
    4. {'POS': 'VERB', 'OP': '+'},
    5. {'POS': 'ADV', 'OP': '*'},
    6. {'POS': 'ADJ', 'OP': '+'},
    7.]
\end{verbatim}

The numbered lines respectively describe: 1) a wildcard representing any token; 2) an optional auxiliary such as "is", "was", "could", or "should"; 3) an optional negation such as "not"; 4) at least one verb such as "trend", "trending", or "react"; 5) none or multiple adverbs such as "significantly"; 6) at least one adjective such as "worse" or "better". Thus, an example VP matching these rules could present as "\textit{[entity] is trending significantly worse}".

The combination of the sentiment bias vector and the sentiment-based query expansion will hereafter be referred to as \textit{Sentiment Biases (SB)}.

%% file: Sections/ResultsAndDiscussion.tex
\begin{table*}
    \centering
    \begin{tabular}{l|l|c|r|r|r|r}
    
    
     \multicolumn{1}{c|}{\(\boldsymbol\alpha\)} &
     \multicolumn{1}{c|}{\(\boldsymbol\beta\)} &
     \multicolumn{1}{c|}{\textbf{Experiments}} &
     \multicolumn{1}{c|}{\small\textbf{R-1}} &
     \multicolumn{1}{c|}{\small\textbf{R-2}} &
     \multicolumn{1}{c|}{\small\textbf{R-L}} &
     \multicolumn{1}{c}{\small\textbf{R-SU4}}\\
    \hline
    
    - & - & \small{Upper bound} & 72.86 & 48.60 & 72.05 & 49.63\\ 
    \hline
    
    - & - & \small{BQ\(\rightarrow\)MMR} & 25.13 & 8.59 & - & 10.29\\ 
    
    - & - & \small{BQ\(\rightarrow\)QuerySum} & 27.03 & 12.03 & - & 12.86\\ 
    \cline{3-3}
    
    0.85 & - & \multirow{2}{*}{\small{BQ\(\rightarrow\)BTR}} & 31.98 & 16.91 & 28.61 & 16.69\\ 
    
    0.1 & - & & \textbf{34.15} & \textbf{17.50} & \textbf{30.35} & \textbf{17.41}\\ 
    \hline
    
    0.85 & - & \multirow{2}{*}{\small{FRW-MPB2\(\rightarrow\)BTR}} & 32.73 & 17.48 & 29.06 & 17.37\\
    
    0.1 & - & & \textbf{41.67} & \textbf{24.50} & \textbf{37.69} & \textbf{24.35}\\ 
    \cline{3-3}
    
    0.85 & - & \multirow{2}{*}{\small{FRP-BTR\(\rightarrow\)BTR}} & 33.20 & 17.70 & 29.48 & 17.48\\ 
    
    0.1 & - & & 37.79 & 21.18 & 34.21 & 20.77\\ 
    \cline{3-3}
    
    0.85 & - & \multirow{2}{*}{\small{FRP-MPB2\(\rightarrow\)BTR}} & 31.97 & 17.12 & 28.50 & 16.78\\ 
    
    0.1 & - & & 38.57 & 22.32 & 34.98 & 21.76\\ 
    \hline
    
    0.1 & 0.1 & \multirow{3}{*}{\small{ERT\(\rightarrow\)MBTR}} & \textbf{45.51} & \textbf{28.22} & \textbf{41.61} & \textbf{28.11}\\ 
    
    0 & 0.1 & & 44.21 & 27.02 & 40.03 & 26.96\\
    
    0.1 & 0 & & 44.82 & 27.84 & 41.01 & 27.67\\ 
    \hline
    
    0.1 & 0.1 & \multirow{4}{*}{\small{SB\(\rightarrow\)MBTR}} & 43.58 & 25.45 & 39.10 & 25.36\\
    
    
    0.1 & 0 & & 42.51 & 24.89 & 38.44 & 25.01\\
    
    0.1 & 0.2 & & \textbf{44.11} & \textbf{25.77} & \textbf{39.58} & \textbf{25.64}\\
    
    0 & 0.2 & & 43.42 & 25.18 & 38.93 & 25.02\\
    
    \end{tabular}
    \caption[ROUGE scores of varied query combinations]{
    ROUGE scores of our \ref{MBTR_ICR_QE} and \ref{SAMBTR} experiments. We use the left-hand side of \(\boldsymbol\rightarrow\) to denote the query inputs. The \textit{upper bound} is computed by selecting the source sentence with the highest ROUGE-SU4 score (\ref{ROUGE}). Bold font denotes each subtable's top ROUGE variant score. We use "Why did \{queried product\} receive \{positive, negative\} feedback" as a Baseline Query (\textit{BQ}). In MMR, sentence similarity is computed with spaCy's \textit{en\_core\_web\_lg} model. We use the same SBERT encoder (\ref{MBTR_ICR_QE}) for BTR and MBTR.
    } 
    \label{tab:results}
\end{table*}

Table \ref{tab:results} presents ROUGE scores of experiments partitioned across the following list of subtables:

\begin{enumerate}
    \item The upper bound expresses the maximum achievable scores given that the references are abstractive summaries.
    
    \item MMR, QuerySum, and BTR are used as baseline MDQFES models for comparison. BTR\(|_{\alpha=0.1}\) performs best among baselines across all reported ROUGE variants.
    
    \item Each query expansion from ERT (\ref{MBTR_ICR_QE}) is tested individually on BTR\(|_{\alpha=\{0.1, 0.85\}}\). The FRW-MPB2 query performs best across all reported ROUGE variants.
    
    \item MBTR\(|_{\alpha=\{0, 0.1\} \times \beta=\{0, 0.1\}}\) is tested with ERT as input. MBTR\(|_{\alpha=0.1, \beta=0.1}\) performs best across all reported ROUGE variants. It also outperforms BTR with each ERT query (subtable 3), thus demonstrating the benefit of CBFS; this holds even with ablation of the ICR component (\ref{ICR}) with MBTR\(|_{\alpha=0.1, \beta=0}\).
    
    \item MBTR\(|_{\alpha=\{0, 0.1\} \times \beta=\{0, 0.1, 0.2\}}\) is tested with SB (\ref{SAMBTR}) as input. MBTR\(|_{\alpha=0.1, \beta=0.2}\) performs best across all reported ROUGE variants.
\end{enumerate}

Only the best-performing combinations of \(\alpha\) and \(\beta\) are reported, in addition to combinations relevant to ablation studies.

Throughout all BTR and MBTR experiments, we observe that \(\alpha = 0.1\) performs consistently better than \citet{kazemi_biased_2020}'s recommended 0.85 and than the ablation of the centrality component with \(\alpha = 0\). This suggests that the solution space of ESS with short summaries (\ref{ROUGE}) highly prioritizes query focus, without discarding the intra-document salience component since it helps elect the most central sentence among the most bias-relevant.

Dampening ICR performs best at \(\beta = 0.1\) for the ERT experiments and at \(\beta = 0.2\) for the SB experiments. Thus, for the problem space of ESS with short summaries, we recommend \(\beta = 0.1\) when a development set is available for constructing the ERT queries, and \(\beta = 0.2\) with SB otherwise. We interpret ERT's lesser regularization requirement as benefiting from its inherent proximity with the target specificity given its embedded answer signals (\ref{RQF}).

%% file: Sections/Conclusion.tex
We approach the putative linguistic dissonance in the QFS task with the CBFS framework, which we concretize with the MBTR model. We then specifically address our purpose of sentiment explanation by introducing the ESS task and its system comprising sentiment-based biases and query expansions.

We find that the MBTR model significantly outperforms baseline QFES models and the BTR model it extends. In particular, given that we input the same queries individually to BTR, outperforming it substantiates the CBFS claim of favoring desired sentences through multiple query formulations. Our results also indicate that the ESS task is more suitable than QFS when the query involves a known sentiment.

This work is limited by its focus on the problem space of single-sentence reference summaries and by its lack of testing on other ESS datasets. In future works, we plan on adapting ABSA datasets to the ESS task  and on integrating other QFS models into the CBFS framework. Additionally, asymmetric semantic search encoders, such as those we used for query expansion in ESS (\ref{SAMBTR}), might be better suited for the QFES process when the desired summaries are longer than one sentence.